\documentclass[letterpaper, 10 pt, conference]{ieeeconf}

\IEEEoverridecommandlockouts
\usepackage{amsmath,amssymb}
\usepackage{graphicx}
\usepackage{booktabs}
\usepackage{multirow}
\usepackage{subcaption}
\usepackage{float}
\usepackage{algorithm}
\usepackage{algpseudocode}
\usepackage{xcolor}
\usepackage{cite}
\usepackage{url}
\usepackage{hyperref}

\graphicspath{{figures/}}

\title{\LARGE \bf
O-ConNet: Geometry-Aware End-to-End Inference of\\Over-Constrained Spatial Mechanisms
}

\author{Haoyu Sun$^{1}$, Meng Zhao$^{1}$, Tianhao Wang$^{1}$, and Jianxu Wu$^{1*}$%
\thanks{$^{1}$School of Mechanical, Electronic and Control Engineering, Beijing Jiaotong University, Beijing, China.}%
\thanks{$^{*}$Corresponding author: J. Wu (wujx@bjtu.edu.cn).}%
\thanks{This work was supported by the National Natural Science Foundation of China (Grant No. 52575002).}%
}

\begin{document}

\maketitle
\thispagestyle{empty}
\pagestyle{empty}

\begin{abstract}
Deep learning-driven scientific discovery has achieved breakthroughs in microscopic domains such as protein folding and molecular synthesis, yet this paradigm has not been fully examined in macroscopic rigid-body~kinematics.
Spatial over-constrained mechanisms are governed by stringent geometric-algebraic constraints, making their generative design a litmus test for whether AI can truly internalize three-dimensional physical-topological~constraints.
Focusing on rigid over-constrained mechanisms, this paper proposes O-ConNet (Over-Constrained Network): an end-to-end learning framework that directly predicts mechanism structural parameters from only three sparse reachable points while simultaneously reconstructing the complete motion trajectory, without explicitly solving any constraint~equations.
On a self-constructed dataset of 42,860 samples, O-ConNet achieves a Param-MAE of $0.276\pm0.077$ and a Traj-MAE of $0.145\pm0.018$ (mean\,$\pm$\,std over 10 independent runs), outperforming the best sequence-prediction baseline (LSTM-Seq2Seq) by 65.1\% and 88.2\%,~respectively.
These results provide preliminary evidence that end-to-end deep learning is a viable approach to the inverse design of spatial over-constrained~mechanisms.
This work represents an exploratory attempt to understand macroscopic rigid-body geometric constraints in a data-driven manner, offering a feasible methodology for computationally synthesizing high-performance spatial mechanisms that may surpass human~intuition.
\end{abstract}

\section{INTRODUCTION}\label{sec:intro}

The persistent industrial demand for spatial motion precision has made the inverse design of spatial closed-chain mechanisms a recurring engineering bottleneck: a designer can only specify a few key waypoints that the end-effector must pass through, yet must accurately recover structural parameters from a parameter space subject to stringent geometric-algebraic~constraints.
While forward kinematics yields a unique motion trajectory for any given parameter set, the inverse direction must recover structural parameters from only a small number of observed positions, a task that is inherently ill-posed and poses substantial~challenges.
This asymmetry motivates investigating whether a neural network trained solely on data distributions satisfying the constraints can implicitly internalize three-dimensional rigid-body geometric constraints as a generalizable structural representation in latent vector space, thereby supporting end-to-end inverse inference from extremely sparse observations to~precise~parameters.

Spatial over-constrained mechanisms are ideal objects for examining this question: their assemblable parameter sets occupy only an extremely sparse subset of the full parameter space, and a model must distill the intrinsic geometric structure of the constraint manifold from limited samples to generalize reliably on unseen~parameters.
This paper takes the Bennett 4R mechanism---the minimal and most tightly constrained representative among such mechanisms---as the \textbf{Minimum Viable Model}, and proposes O-ConNet with the following main~contributions:
\begin{itemize}
  \item \textbf{Task Framework}: We formulate the inverse design of spatial over-constrained mechanisms as end-to-end learning from three sparse waypoints to structural parameters, with auxiliary trajectory reconstruction enabling implicit internalization of closed-loop geometric~constraints.
  \item \textbf{Technical Design}: We present a network design that exploits trajectory reconstruction as a geometric constraint proxy, with encoding and loss strategies adapted to the sparse-observation, angular-periodic, and multi-task nature of~the~problem.
  \item \textbf{Evaluation Benchmark}: A 42,860-sample Bennett 4R dataset, a dual-metric (parameter + trajectory) evaluation protocol, and three architectural baselines are provided as a~reproducible~testbed.
\end{itemize}

Section~\ref{sec:related} situates this work within existing traditional and learning-based~approaches.
Section~\ref{sec:formulation} formalizes the Bennett mechanism, derives the two-parameter constraint structure that defines the learning task, and describes dataset~construction.
Section~\ref{sec:model} presents the network architecture and loss function, with each design choice tied to a specific challenge identified~above.
Section~\ref{sec:experiments} evaluates these choices through baseline comparisons and ablation~studies.
Section~\ref{sec:conclusion} concludes with a discussion of broader~implications.

\section{RELATED WORK}\label{sec:related}

Traditional methods for mechanism synthesis can be classified into three~categories~\cite{pathak2023historical}.
Analytical methods~\cite{perez2004dual} employ dual quaternions or algebraic elimination to obtain exact closed-form solutions under special configurations; while theoretically rigorous and verifiable, extending them to three-dimensional pose-coupled over-constrained mechanisms poses significant scaling~challenges.
Metaheuristic methods, represented by genetic algorithms and particle swarm optimization~\cite{cabrera2002optimal}, bypass solvability limitations through stochastic sampling and exhibit broad applicability, though convergence speed in high-dimensional parameter spaces remains an~open~issue.
Atlas-based methods~\cite{sun2020synthesis} combine discretized databases with fast retrieval, suited for real-time scenarios, but their coverage precision over continuous parameter domains is bounded by the database~size.
Each category possesses distinct advantages, collectively forming a solid foundation for mechanism synthesis~research.
Kong~\cite{kong2025type} further exemplifies this line of inquiry by performing type synthesis of variable-DOF single-loop spatial mechanisms built upon Bennett 4R and Goldberg 5R loops, confirming that over-constrained closed chains remain a structurally central and actively studied building block in spatial mechanism~design.
These advances establish robust foundations, while manifold complexity motivates exploring complementary~approaches.

The breakthroughs of deep learning in domains such as protein folding~\cite{jumper2021highly} demonstrate that data-driven paradigms can capture deep structural regularities in physical systems. This insight is beginning to extend to macroscopic rigid-body~kinematics.
For planar four-bar mechanisms, Deshpande et al.~\cite{deshpande2019machine} employed machine learning to achieve end-to-end prediction from trajectory images to~parameters.
Lin et al.~\cite{lin2025creative} applied a VAE-based generative synthesis to eight-bar mechanisms using images as the medium, explicitly modeling the trajectory manifold in latent space via an autoencoder to improve generalization to unseen~trajectories.
Jadhav and Farimani~\cite{jadhav2026linkd} introduced an autoregressive diffusion model for conditional generation of planar linkage~mechanisms.
For spatial mechanisms, Deng et al.~\cite{deng2025path} completed path synthesis of RSCR spatial mechanisms through an encoder--decoder~framework.
These works reveal a core insight: \emph{trajectory reconstruction can serve as a proxy task for parameter inversion}, enabling the network to implicitly encode kinematic constraint structures in feature~space.
The assemblable manifold of spatial single-loop over-constrained mechanisms is an extremely sparse subset of the full parameter space, and inverse design simultaneously faces the dual challenges of three-dimensional screw coupling and constraint manifold sparsity. To date, no end-to-end framework has been specifically developed for this class of~mechanisms.

Large-scale generatable datasets form the material basis for data-driven~synthesis.
Li and Chen~\cite{li2017parametrization} proposed parametrization-invariant Fourier descriptors for path~normalization.
Heyrani Nobari et al.~\cite{heyrani2022links} constructed LINKS, a dataset containing one hundred million planar linkage~samples.
On the technical tool front, PointNet~\cite{qi2017pointnet} provides a permutation-invariant point cloud encoding paradigm that naturally suits the processing of unordered sparse spatial~observations.
The Transformer~\cite{vaswani2017attention} offers powerful long-range modeling capability for efficient periodic three-dimensional trajectory~reconstruction.
ResNet-style~\cite{he2016deep} residual shortcut connections provide stable gradient pathways for deep parameter prediction~heads.
This combination of tools has not yet been fully demonstrated in the inverse design of spatial over-constrained mechanisms; the present work constructs O-ConNet at precisely this~intersection.

\section{PROBLEM FORMULATION AND DATASET CONSTRUCTION}\label{sec:formulation}

\subsection{Bennett Mechanism Parameterization}\label{sec:parameterization}

\begin{figure}[t]
  \centering
  \includegraphics[width=0.5\columnwidth]{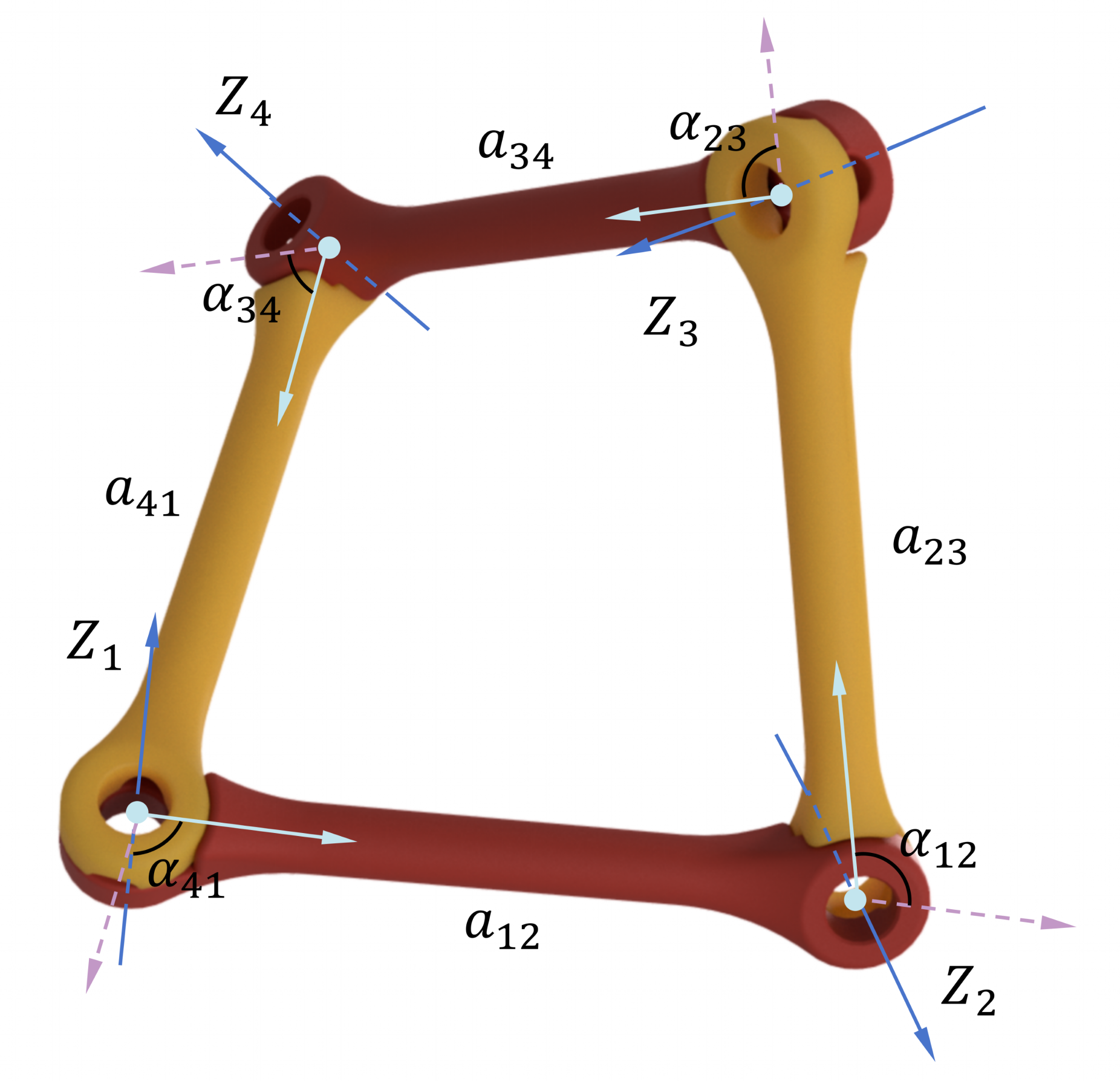}
  \caption{D-H parameterization schematic of the Bennett mechanism.}
  \label{fig:bennett_schematic}
\end{figure}

\begin{figure*}[!t]
  \centering
  \includegraphics[width=\textwidth]{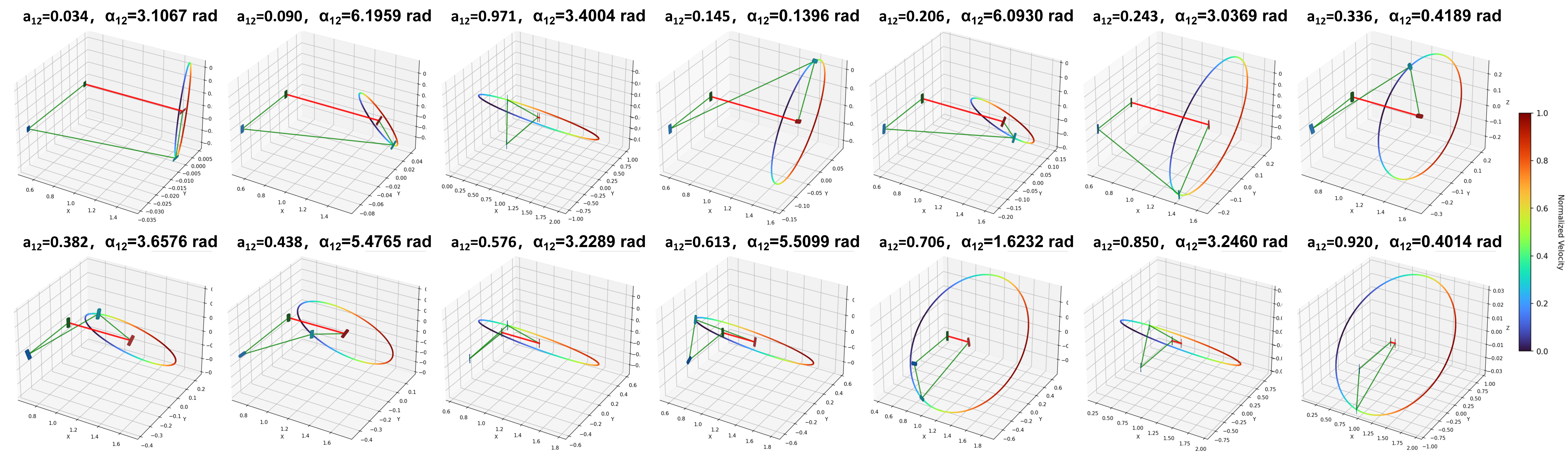}
  \caption{3D trajectory visualization of dataset samples. Each panel displays the complete motion trajectory of a Bennett mechanism's third joint endpoint, with speed magnitude encoded by a plasma colormap.}
  \label{fig:dataset_samples}
\end{figure*}

The Bennett mechanism consists of four links connected end-to-end by revolute joints to form a spatial closed chain~\cite{lee1995kinematic}, as illustrated in Fig.~\ref{fig:bennett_schematic}.
The modified D-H convention is adopted to describe the geometric relationships between adjacent joints: $a_i$ denotes the common normal distance between the two revolute axes bounding the $i$-th link (link length), $\alpha_i$ is the twist angle between adjacent axes, the offset $d_i{\equiv}0$, and $\theta_i$ is the joint variable (joint angle) of joint $i$. Specifically, the parameters connecting joint $i$ to joint $i{+}1$ are denoted $a_i$ and $\alpha_i$ (e.g., $\alpha_{12}$ denotes the twist angle of the link between joints~1~and~2).
The homogeneous transformation matrix at each joint~is:
\begin{equation}\label{eq:dh}
  T_i(\theta_i) = \begin{bmatrix}
    c_{\theta_i} & -s_{\theta_i}c_{\alpha_i} &  s_{\theta_i}s_{\alpha_i} & a_i c_{\theta_i} \\
    s_{\theta_i} &  c_{\theta_i}c_{\alpha_i} & -c_{\theta_i}s_{\alpha_i} & a_i s_{\theta_i} \\
    0            &  s_{\alpha_i}             &  c_{\alpha_i}             & 0               \\
    0            &  0                        &  0                        & 1
  \end{bmatrix}
\end{equation}
where $c_x{=}\cos x$ and $s_x{=}\sin x$. The concatenation of four transformations must satisfy the spatial closure~constraint:
\begin{equation}\label{eq:loop}
  T_1(\theta_1)\,T_2(\theta_2)\,T_3(\theta_3)\,T_4(\theta_4) = I
\end{equation}
Given the driving angle $\theta_1$, this system of equations uniquely determines the follower angles $(\theta_2,\theta_3,\theta_4)$; sweeping $\theta_1\in[0,2\pi)$ traces the trajectory of the third joint~endpoint.

For a general spatial four-bar mechanism, the link lengths and twist angles of the four links yield eight independent D-H parameters $(a_{12},a_{23},a_{34},a_{41},\alpha_{12},\alpha_{23},\alpha_{34},\alpha_{41})$.
However, the necessary and sufficient condition for~(\ref{eq:loop}) to admit real solutions over the entire $[0,2\pi)$ imposes stringent geometric constraints on these eight parameters, ultimately compressing the degrees of freedom to only \textbf{two}. Two sequential reductions achieve~this.

Opposite links must satisfy the equal-length and equal-twist~symmetry:
\begin{equation}\label{eq:sym}
  a_{12}=a_{34},\quad a_{23}=a_{41},\quad \alpha_{12}=\alpha_{34},\quad \alpha_{23}=\alpha_{41}
\end{equation}
This symmetry folds eight parameters into four independent quantities $(a_{12},a_{23},\alpha_{12},\alpha_{23})$. On this basis, adjacent links must further satisfy the proportionality~relation:
\begin{equation}\label{eq:bennett}
  \frac{\sin\alpha_{12}}{a_{12}} = \frac{\sin\alpha_{23}}{a_{23}}
\end{equation}
Once $(a_{12},a_{23},\alpha_{12})$ are specified, the twist angle $\alpha_{23}$ is therefore uniquely determined, reducing the count of independent parameters to~three.
Noting further that the kinematic behavior of the mechanism depends only on the relative proportions of the links, not on the absolute scale, we~impose:
\begin{equation}\label{eq:scale}
  a_{12} + a_{23} = 1
\end{equation}
to eliminate the redundant scale degree of freedom, whereby $a_{23}=1-a_{12}$ becomes a derived~quantity.
The independent parameters required to fully describe a Bennett mechanism thus reduce to exactly~two:
\begin{equation}\label{eq:params}
  a_{12}\in(0,1),\qquad \alpha_{12}\in(0,2\pi)
\end{equation}

It is precisely based on these constraints that the inverse design of the Bennett mechanism essentially degenerates into a two-dimensional search problem, and the learning objective of O-ConNet is accordingly~determined.

\subsection{Input Representation and Learning Task}\label{sec:task}

The typical starting point for mechanism inverse design is: the designer specifies a small number of \textbf{desired waypoints} that the end-effector must pass through precisely, and the goal is to recover the mechanism structural parameters that realize these configurations. In this paper, each desired waypoint is represented as a 7-dimensional kinematic state~vector:
\begin{equation}\label{eq:feat}
  \mathbf{x}_i = (\,\underbrace{p_x,p_y,p_z}_{\text{position}},\;\underbrace{v_x,v_y,v_z}_{\text{velocity}},\;\underbrace{r}_{\text{speed ratio}}\,)\in\mathbb{R}^7
\end{equation}
where the three-dimensional position comes from the third joint endpoint coordinates, the velocity vector is estimated by central differencing, and the normalized speed ratio $r{=}|\mathbf{v}_i|/\max_j|\mathbf{v}_j|\in[0,1]$ eliminates the influence of absolute velocity magnitude. The three components collectively provide complementary descriptions of configuration, motion direction, and speed~distribution.

From $N_{\text{in}}{=}3$ desired waypoints $\{\mathbf{x}_i\}_{i=1}^{3}$ (uniformly sampled from the complete trajectory), the core learning objective is to recover the mechanism parameters $(\hat{a}_{12},\hat{\alpha}_{12})$ from extremely sparse~input.
The complete trajectory for each training sample originates from the forward kinematics solver operating over 360 uniformly spaced driving angles (see Algorithm~\ref{alg:dataset}). From this 360-frame sequence, 64 representative frames are uniformly subsampled to constitute the dense ground-truth trajectory, and the three input waypoints are in turn drawn at equal intervals from this 64-point~set.
Directly regressing parameters from three discrete configurations is highly underdetermined in an information-theoretic sense; therefore, the following strategy is~adopted:
The network first reconstructs a complete 64-point densely sampled trajectory $\hat{\mathcal{T}}\in\mathbb{R}^{64\times3}$ from the three sparse points, then feeds both the reconstructed trajectory and the original three observations into a joint pathway that collaboratively infers the mechanism structural~parameters.
By routing gradients through trajectory reconstruction before reaching the parameter head, the closed-loop constraint structure implicit in parameter space becomes encoded in the shared latent representation, while the combined trajectory-and-observation pathway equips the parameter head with richer geometric and kinematic discriminative signals than either source alone could~provide.

Although the encoder handles arbitrary $N_{\text{in}}$, this paper universally fixes $N_{\text{in}}{=}3$.
The choice is grounded in geometric minimality: two spatial points constrain at most an arc segment and cannot disambiguate distinct closed-loop configurations, whereas three non-collinear points constitute the minimal non-degenerate specification for a spatial curve, placing $N_{\text{in}}{=}3$ at the lowest cardinality at which the inverse problem is~well-posed.
Adopting this most information-sparse setting further ensures that successful parameter recovery reflects genuine encoding of the constraint manifold rather than reliance on input~redundancy.

\subsection{Dataset Construction and Normalization}\label{sec:dataset}

\begin{algorithm}[t]
\caption{Bennett 4R Dataset Generation}
\label{alg:dataset}
\begin{algorithmic}[1]
\Require $N{=}360$: frames per sample;
         \Statex\hspace{2.6em}$\theta_1$ sampled uniformly on $[0,2\pi)$
\Require $\epsilon{=}10^{-6}$: LM solver convergence tolerance
\Require $f_c{=}72$: low-pass cutoff frequency;
         \Statex\hspace{2.6em}harmonics ${\le}f_c$ retained
\Require $\tau{=}3\sigma_{\mathcal{T}}$: inter-frame jump threshold;
         \Statex\hspace{2.6em}$\sigma_{\mathcal{T}} = \mathrm{std}_k\|\mathcal{T}_{k+1}-\mathcal{T}_k\|$
\Ensure  $\mathcal{D} = \bigl\{(a_{12}^{(i)},\,\alpha_{12}^{(i)},\,\mathcal{T}_i^*)\bigr\}$:
         valid samples with filtered trajectories
\Statex\rule{\linewidth}{0.3pt}
\State Build candidate grid:
       \Statex\hspace{2em}$a_{12}\!\in\![0.02,0.48]\cup[0.52,0.98]$
       \Statex\hspace{2em}$\alpha_{12}\!\in\![5^\circ,178^\circ]\cup[185^\circ,355^\circ]$
\For{each candidate $(a_{12},\,\alpha_{12})$}
  \State $a_{23} \leftarrow 1 - a_{12}$
  \State $\sin\alpha_{23} \leftarrow \dfrac{a_{23}}{a_{12}}\sin\alpha_{12}$
         \Comment{Bennett constraint~(\ref{eq:bennett})}
  \Statex\hspace{1.5em}\textbf{Gate 1} \textemdash\ real-solution check
  \State \textbf{if} $|\sin\alpha_{23}| > 1$\; \textbf{then skip}
  \State $(\theta_{2,k},\theta_{3,k},\theta_{4,k})
           \leftarrow \mathrm{LM}\!\left(\prod_{j=1}^{4}T_j(\theta_{j,k}){=}I;\;\epsilon\right)$,\;
         $k{=}0,\ldots,N{-}1$
  \State $n_\mathrm{ok} \leftarrow \#\{\text{converged frames}\}$
  \Statex\hspace{1.5em}\textbf{Gate 2} \textemdash\ assemblability check
  \State \textbf{if} $n_\mathrm{ok} < 0.8N$\; \textbf{then skip}
  \State $\mathcal{T}^* \leftarrow \mathrm{DFT\text{-}LPF}(\mathcal{T},\,f_c)$
  \Statex\hfill$\triangleright$\enspace suppress numerical noise
  \Statex\hspace{1.5em}\textbf{Gate 3} \textemdash\ branch-jump check
  \State \textbf{if} $\max_k\|\mathcal{T}^*_{k+1}-\mathcal{T}^*_k\| > \tau$\; \textbf{then skip}
  \State $\mathcal{D} \leftarrow \mathcal{D} \cup \{(a_{12},\,\alpha_{12},\,\mathcal{T}^*)\}$
\EndFor
        \hfill$\bigl(|\mathcal{D}|{=}42{,}860$;\; ${\approx}62\%$ pass rate$\bigr)$
\State \Return $\mathcal{D}$
\end{algorithmic}
\end{algorithm}

\begin{algorithm}[t]
\caption{Two-Stage Coordinate Normalization}
\label{alg:normalize}
\begin{algorithmic}[1]
\Require $\mathcal{D}$: raw dataset from Algorithm~\ref{alg:dataset}
\Ensure  $\tilde{\mathcal{D}}$: normalized dataset with coordinates in $[-1,1]$
\Statex\rule{\linewidth}{0.3pt}
\For{each sample $(a_{12},\,\alpha_{12},\,\mathcal{T}^*)\in\mathcal{D}$}
  \State $\kappa \leftarrow 1/a_{23}$
         \hfill$\triangleright$\enspace $a_{23}=1{-}a_{12}$ prior to scaling
  \State $(a_{12},\,\mathcal{T}^*) \leftarrow \kappa\cdot(a_{12},\,\mathcal{T}^*)$
  \Comment{Stage~1: set $a_{23}{=}1.0$}
\EndFor
\State $c_{99}\leftarrow\mathrm{percentile}_{99}\bigl(\{\|\mathcal{T}^*\|\}\bigr)$
       \hfill$\triangleright$\enspace computed as $29.44$ on training split
\For{each sample $\in\mathcal{D}$}
  \State $(a_{12},\,\mathcal{T}^*) \leftarrow (a_{12},\,\mathcal{T}^*)\,/\,c_{99}$
  \Statex\hfill$\triangleright$\enspace Stage~2: global percentile scaling
\EndFor
\State \Return $\tilde{\mathcal{D}}$
       \hfill$\bigl(a_{12}\!\in\!(0,1.66)$;\; 99\% of coords in $[-1,1]\bigr)$
\end{algorithmic}
\end{algorithm}

\subsubsection{Generation Pipeline}
The central challenge in constructing a high-quality dataset is that the Bennett constraint admits real solutions on only an extremely sparse subset of the parameter space, and the numerical forward kinematics solver suffers from ill-conditioned Jacobians in certain parameter~regions.
The complete generation pipeline is given in Algorithm~\ref{alg:dataset}; the design choices at each stage are explained~below.

The non-uniform screening of the parameter space deliberately avoids three types of~regions.
When $a_{12}$ approaches 0 or 1, the opposite link length ratio becomes extreme, causing the Jacobian condition number of the LM solver to increase~sharply.
When $a_{12}\approx 0.5$ (i.e., $a_{12}=a_{23}$), the constraint equations exhibit a symmetric solution structure, and the numerical solver is prone to branch~switching.
When $\alpha_{12}\approx 0^\circ$, the spatial mechanism degenerates into a planar four-bar; when $\alpha_{12}\approx 180^\circ$, adjacent link axes are nearly anti-parallel, and the valid solutions of the Bennett constraint become drastically sparse in this~neighborhood.
Accordingly, $a_{12}$ is sampled from $[0.02,0.48]\cup[0.52,0.98]$ and $\alpha_{12}$ from $[5^\circ,178^\circ]\cup[185^\circ,355^\circ]$.

The three gates serve distinct~purposes.
Gate~1 analytically verifies $|\sin\alpha_{23}|\le 1$ using the Bennett closure constraint, filtering out candidates with no physically real solutions before simulation at negligible~cost.
Gate~2 uses a success rate threshold of 80\% rather than 100\% to tolerate a small number of anomalous frames at trajectory endpoints, avoiding excessive rejection of assemblable~samples.
Gate~3 applies low-pass filtering to address occasional branch jumps between adjacent frames in the LM solver, and the $3\sigma$ test discards samples that still exhibit large inter-frame jumps after~filtering.
The three gates collectively retain approximately 62\% of candidates, yielding 42,860 valid~samples.

Fig.~\ref{fig:dataset_samples} displays 14 randomly selected samples from the dataset, intuitively illustrating the diversity of trajectories generated by Bennett mechanisms at different locations in parameter~space.

\subsubsection{Two-Stage Normalization}
Without normalization, trajectory magnitudes conflate mechanism geometry with the sample-varying $a_{23}$, and coordinates routinely exceed the saturation threshold of Sigmoid and Tanh near $\pm3$. Algorithm~\ref{alg:normalize} treats both effects in sequence: Stage~1 fixes $a_{23}{=}1$ per sample to isolate shape from scale, while Stage~2 applies a global 99th-percentile divisor that confines 99\% of coordinates to $[-1,1]$ and restores healthy gradient flow. The renormalized $a_{12}$ range of $(0,1.66)$ motivates $A_{\mathrm{max}}{=}2.0$ in~(\ref{eq:a12}); all metrics in Section~\ref{sec:experiments} are reported in this coordinate~system.

\section{MODEL}\label{sec:model}

\begin{figure*}[!t]
  \centering
  \includegraphics[width=0.98\textwidth]{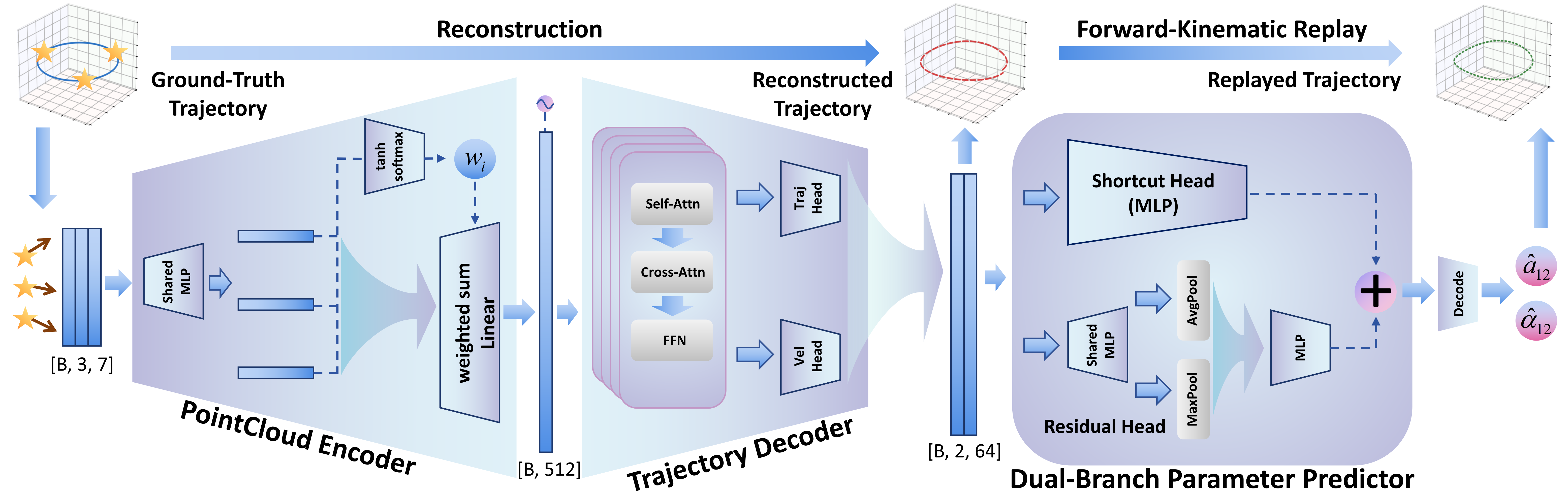}
  \caption{Complete architecture of O-ConNet. On the left, a permutation-invariant kinematic state encoder processes 3 sparse observation points through a shared-weight MLP and learnable attention pooling (AttnPool), compressing them into a global latent vector $\mathbf{z}\in\mathbb{R}^{512}$. In the center, a Transformer decoder expands $\mathbf{z}$ into 64-point trajectory coordinates $\hat{\mathbf{T}}$ and a velocity field $\hat{\mathbf{V}}$. On the right, the parameter prediction head adopts a ResNet-style shortcut-residual dual-path structure, where the shortcut branch maps $\mathbf{z}$ directly to coarse logits~$\mathbf{s}$ (supervised by $\mathcal{L}_{\mathrm{aux}}$) and the residual branch reads the reconstructed trajectory to produce a zero-initialized refinement~$\mathbf{r}$; the merged output is decoded via Sigmoid and atan2 into $(\hat{a}_{12},\hat{\alpha}_{12})$. In all trajectory visualizations throughout this paper, blue solid lines denote ground truth (GT), red dashed lines show the network prediction (Pred), green dotted lines represent the forward kinematic reconstruction from predicted parameters (Kin), and orange stars mark the input observation points.}
  \label{fig:network}
\end{figure*}

O-ConNet adopts a ``shared encoder--dual-branch'' architecture (Fig.~\ref{fig:network}): a permutation-invariant kinematic state encoder compresses $N_{\text{in}}{=}3$ sparse observations into a 512-dimensional latent vector $\mathbf{z}$; a Transformer decoder expands $\mathbf{z}$ into a 64-point trajectory and synchronized velocity field; and a ResNet-style shortcut--residual parameter head produces the final predictions. Routing parameter gradients through both the encoder and the reconstructed trajectory compels the shared representation to internalize kinematic structure discriminative for inverse~design.

\subsection{Network Architecture}\label{sec:arch}

Let the $N_{\text{in}}$ sparse input points be $\{\mathbf{x}_i\}_{i=1}^{N_{\text{in}}}$, with $\mathbf{x}_i=(\mathbf{p}_i,\mathbf{v}_i,r_i)\in\mathbb{R}^7$, where $\mathbf{p}_i\in\mathbb{R}^3$ is the 3D coordinate, $\mathbf{v}_i\in\mathbb{R}^3$ is the velocity vector, and $r_i=|\mathbf{v}_i|/\max_j|\mathbf{v}_j|$ is the normalized speed~ratio.

\textbf{Permutation-Invariant Kinematic State Encoder.}
The sparse observations $\{\mathbf{x}_i\}$ lack global temporal alignment and simultaneously carry multi-modal information including position and velocity, precluding any assumption of inherent ordering. To this end, an Unordered Set Mapping Mechanism is designed: drawing on the unordered set processing paradigm of PointNet~\cite{qi2017pointnet}, each kinematic state is independently decoupled through shared-weight per-point high-dimensional projection layers (MLP $7\to64\to128\to256$, with LayerNorm, GELU activation, and Dropout at each layer), followed by \textbf{learnable attention pooling (AttnPool)} for weighted aggregation of the global geometric~representation:
\begin{equation}\label{eq:encoder}
  \mathbf{z} = \mathrm{Proj}\!\left(\textstyle\sum_i w_i\,\mathbf{f}_i\right)\in\mathbb{R}^{512},\quad
  w_i = \frac{\exp(g(\mathbf{f}_i))}{\sum_j\exp(g(\mathbf{f}_j))}
\end{equation}
where $g$ is a two-layer MLP ($256\!\to\!128\!\to\!1$, Tanh activation) and $w_i$ assigns a learnable global participation weight to each observation point. The aggregated feature is projected (Linear $256\!\to\!512$), followed by LayerNorm and GELU, to output a permutation-invariant latent vector $\mathbf{z}$ serving as the global condition for the decoder. Compared to MaxPool's per-dimension discrete argmax selection, attention weights blend contributions from all $N_{\text{in}}$ observation points in a continuously differentiable manner, yielding more thorough gradient flow and richer feature expressiveness under the extremely sparse $N_{\text{in}}{=}3$~setting.

\textbf{Decoder.}
The decoder expands $\mathbf{z}$ into $L{=}64$-point sequential output. A 4-layer Pre-LN Transformer Decoder~\cite{vaswani2017attention} (8-head attention, hidden dimension 256, FFN dimension 1024) takes $\mathbf{z}$ projected through a linear layer as the cross-attention memory, and uses $L$ \textbf{fixed sinusoidal positional encodings} directly as query vectors with no additional learnable query parameters. At the decoder output, two structurally identical independent linear heads produce the trajectory coordinates $\hat{\mathbf{T}}\in\mathbb{R}^{64\times3}$ and velocity vectors $\hat{\mathbf{V}}\in\mathbb{R}^{64\times3}$, respectively (each: LayerNorm $\to$ Linear($256\to128$) $\to$ GELU $\to$ Dropout $\to$ Linear($128\to3$)).

\textbf{Parameter Prediction Head.}
Parameter regression employs a ResNet-style~\cite{he2016deep} shortcut--residual dual-branch structure (see Fig.~\ref{fig:network}, right), balancing prediction accuracy and training~stability.

\emph{Shortcut Branch.} This branch directly predicts coarse parameter logits $\mathbf{s}\in\mathbb{R}^3$ from the encoder latent vector $\mathbf{z}$ through a three-layer MLP ($512\!\to\!256\!\to\!128\!\to\!3$, with LayerNorm, GELU, and Dropout($0.1$) after the first hidden layer), providing a stable global signal during early~training.

\emph{Residual Branch.} Taking reconstructed points $[\hat{\mathbf{p}}_j;\hat{\mathbf{v}}_j]\in\mathbb{R}^6$ as input, this branch again invokes the Unordered Set Mapping Mechanism (MLP $6\!\to\!64\!\to\!128\!\to\!256$, max+avg dual-path pooling aggregation, MLP $256\!\to\!128\!\to\!64$) to extract a refinement residual $\mathbf{r}\in\mathbb{R}^3$ from the reconstructed trajectory. The residual head weights are \textbf{zero-initialized}, making the training start equivalent to the shortcut branch alone. The two paths merge to decode the final~parameters:
\begin{equation}\label{eq:merge}
  \mathbf{m} = \mathbf{s} + \mathbf{r},\qquad
  \mathbf{m} = [m_a,\; m_{\sin},\; m_{\cos}]
\end{equation}
\begin{equation}\label{eq:a12}
  \hat{a}_{12} = \sigma(m_a)\cdot A_{\max} \in (0,\;A_{\max}),\quad A_{\max}=2.0
\end{equation}
\begin{equation}\label{eq:alpha12}
  \hat{\alpha}_{12}
  = \mathrm{atan2}(m_{\sin},\,m_{\cos}) \bmod 2\pi \;\in [0,\,2\pi)
\end{equation}
$a_{12}$ is mapped to $(0,A_{\max})$ via Sigmoid; $A_{\max}{=}2.0$ rather than 1.0 because the actual numerical range of $a_{12}$ after data processing extends to $(0,\,1.66)$, and $A_{\max}{=}2.0$ provides sufficient headroom for the Sigmoid output while preventing predicted values from being clipped to the boundary. $\alpha_{12}$ is represented via the $\sin/\cos$ dual-channel to circumvent angular wrap-around discontinuity, recovered through~$\mathrm{atan2}$.

To further strengthen the independent learning capacity of the shortcut branch, an \textbf{auxiliary loss} $\mathcal{L}_{\mathrm{aux}}$ (\S\ref{sec:training}) is introduced to directly supervise the shortcut branch's parameter predictions $\hat{a}_{12}^{(\mathrm{sc})},\hat{\alpha}_{12}^{(\mathrm{sc})}$, ensuring baseline accuracy even without residual~correction.

\subsection{Loss Function}\label{sec:training}

Uncertainty-adaptive multi-task weighting is adopted, with a learnable log-variance $s_i=\log\sigma_i^2$ for each of the trajectory, velocity, and parameter sub-tasks, automatically balancing dimensional disparities among tasks. The auxiliary branch is supervised with a~fixed~weight:
\begin{equation}\label{eq:loss_total}
  \mathcal{L} = \sum_{i\in\{t,v,p\}}
    \left(\frac{1}{2}e^{-s_i}\mathcal{L}_i + \frac{1}{2}s_i\right)
    \;+\; \lambda_{\mathrm{aux}}\,\mathcal{L}_{\mathrm{aux}}
\end{equation}
where $s_t,s_v$ are initialized to 0 and $s_p$ is initialized to $-2$ (assigning a higher initial precision weight to the parameter~task).

The trajectory loss is $\mathcal{L}_t = \mathrm{MSE}(\hat{\mathbf{T}}, \mathbf{T})$. The velocity loss decouples direction and~magnitude:
\begin{equation}\label{eq:vel_loss}
  \mathcal{L}_v = \left(1 - \overline{\cos}(\hat{\mathbf{V}}, \mathbf{V})\right)
    + \mathrm{L1}\!\left(\log|\hat{\mathbf{V}}|,\, \log|\mathbf{V}|\right)
\end{equation}
Here $\overline{\cos}(\hat{\mathbf{V}},\mathbf{V}) = \frac{1}{LB}\sum_{b,l}\cos(\hat{\mathbf{v}}_{b,l},\mathbf{v}_{b,l})$ is the mean cosine similarity averaged over all $L{=}64$ frames and $B$ batch samples, with $\cos(\mathbf{u},\mathbf{v})\triangleq\mathbf{u}\cdot\mathbf{v}/(\|\mathbf{u}\|\,\|\mathbf{v}\|)$, and $\log|\cdot|$ is applied element-wise to per-frame velocity~norms.
The parameter loss consists of four terms: MSE and \textbf{log-space L1} (handling the right-skewed distribution) for $a_{12}$, and periodicity-aware Huber ($\beta{=}0.1$) and \textbf{$\sin/\cos$ direct L2 supervision} for~$\alpha_{12}$:
\begin{equation}\label{eq:param_loss}
\begin{split}
  \mathcal{L}_p
    &= \underbrace{%
        \mathrm{MSE}(\hat{a}_{12},a_{12})
        + 0.5\,\mathrm{L1}\!\left(\log\hat{a}_{12},\,\log a_{12}\right)
      }_{\text{$a_{12}$ loss}} \\[6pt]
    &\quad+ \underbrace{%
        \begin{array}{c}
          \displaystyle\mathrm{SmoothL1}\!\left(\frac{\Delta\alpha_{12}}{2\pi},\,0;\beta{=}0.1\right)\\[14pt]
          \displaystyle+\;\lambda_{\mathrm{sc}}\!\left(\|\hat{s}-s^*\|^2+\|\hat{c}-c^*\|^2\right)
        \end{array}
      }_{\text{$\alpha_{12}$ loss}}
\end{split}
\end{equation}
\begin{equation}\label{eq:alpha_wrap}
  \Delta\alpha_{12} = \bigl((\hat{\alpha}_{12}-\alpha_{12}+\pi)\bmod 2\pi\bigr)-\pi
\end{equation}
where $\hat{s}=m_{\sin},\hat{c}=m_{\cos}$ are the merged $\sin/\cos$ logits and $s^*=\sin\alpha_{12}^*,c^*=\cos\alpha_{12}^*$ are the ground truth. The $\sin/\cos$ direct supervision bypasses the non-differentiable points of $\mathrm{atan2}$, providing smooth continuous gradient signals, and significantly improves $\alpha_{12}$ accuracy (see Table~\ref{tab:all_results}). The log-space L1 loss $\mathrm{L1}(\log\hat{a}_{12},\log a_{12})$ is more sensitive to small $a_{12}$ values, mitigating the problem of large values dominating the gradient under the right-skewed data~distribution.

The auxiliary loss $\mathcal{L}_{\mathrm{aux}}$ supervises the shortcut branch's independent predictions $\hat{a}_{12}^{(\mathrm{sc})},\hat{\alpha}_{12}^{(\mathrm{sc})}$:
\begin{equation}\label{eq:aux_loss}
  \mathcal{L}_{\mathrm{aux}} = \mathrm{MSE}(\hat{a}_{12}^{(\mathrm{sc})},a_{12})
    + \mathrm{SmoothL1}\!\left(\frac{\Delta\alpha_{12}^{(\mathrm{sc})}}{2\pi},\,0;\beta{=}0.1\right)
\end{equation}
Default hyperparameters are $\lambda_{\mathrm{aux}}{=}10$ and $\lambda_{\mathrm{sc}}{=}1$. The $s_i$ are jointly optimized with the model~parameters.

\section{EXPERIMENTS}\label{sec:experiments}

\begin{figure}[!t]
  \centering
  \includegraphics[width=\columnwidth]{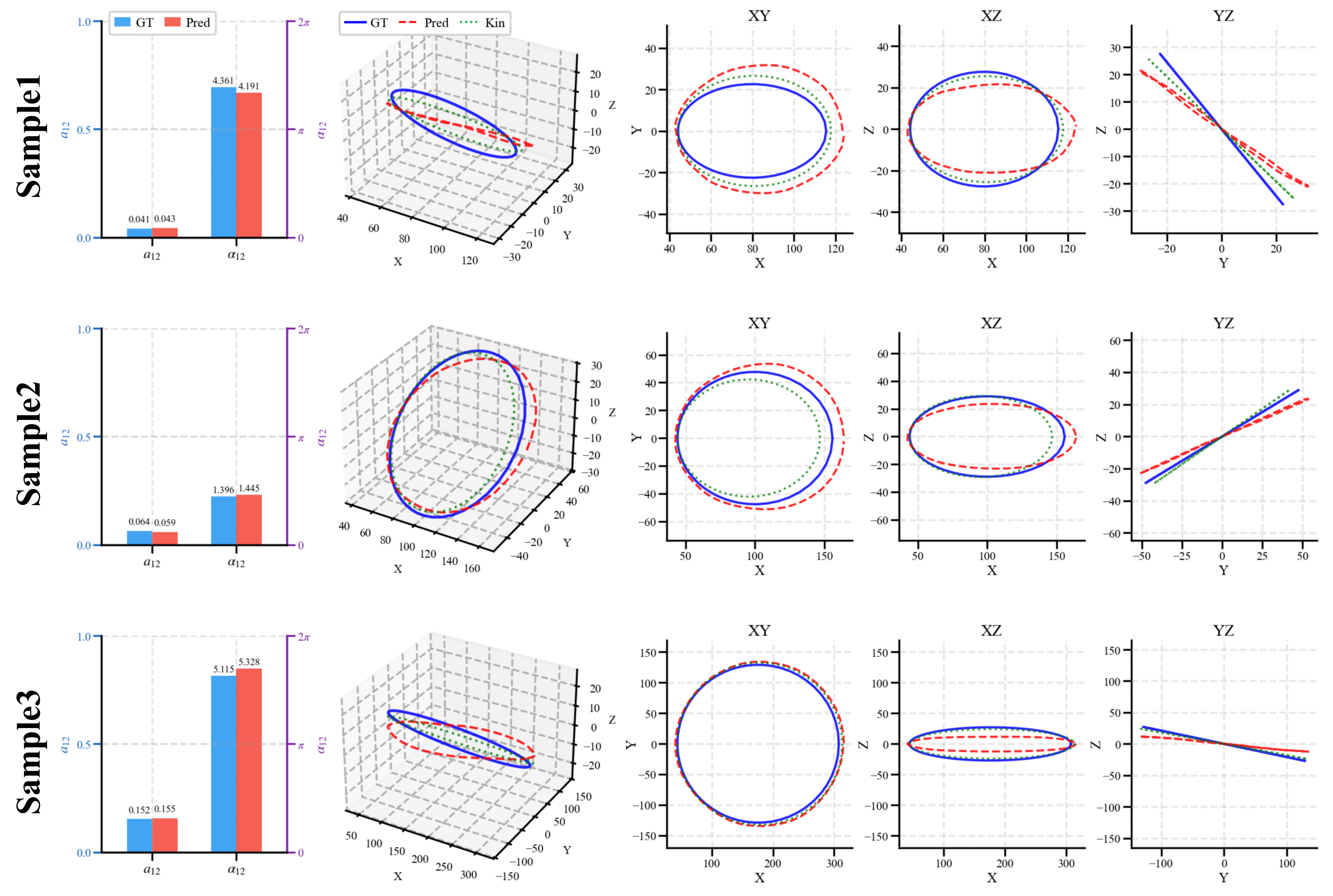}
  \caption{Trajectory prediction and kinematic verification comparison for three representative validation samples. Each row shows one sample; from left to right: 3D view and XY/XZ/YZ projections. Color coding follows Fig.~\ref{fig:network}.}
  \label{fig:traj_vis}
\end{figure}

\begin{figure}[!t]
  \centering
  \includegraphics[width=\columnwidth]{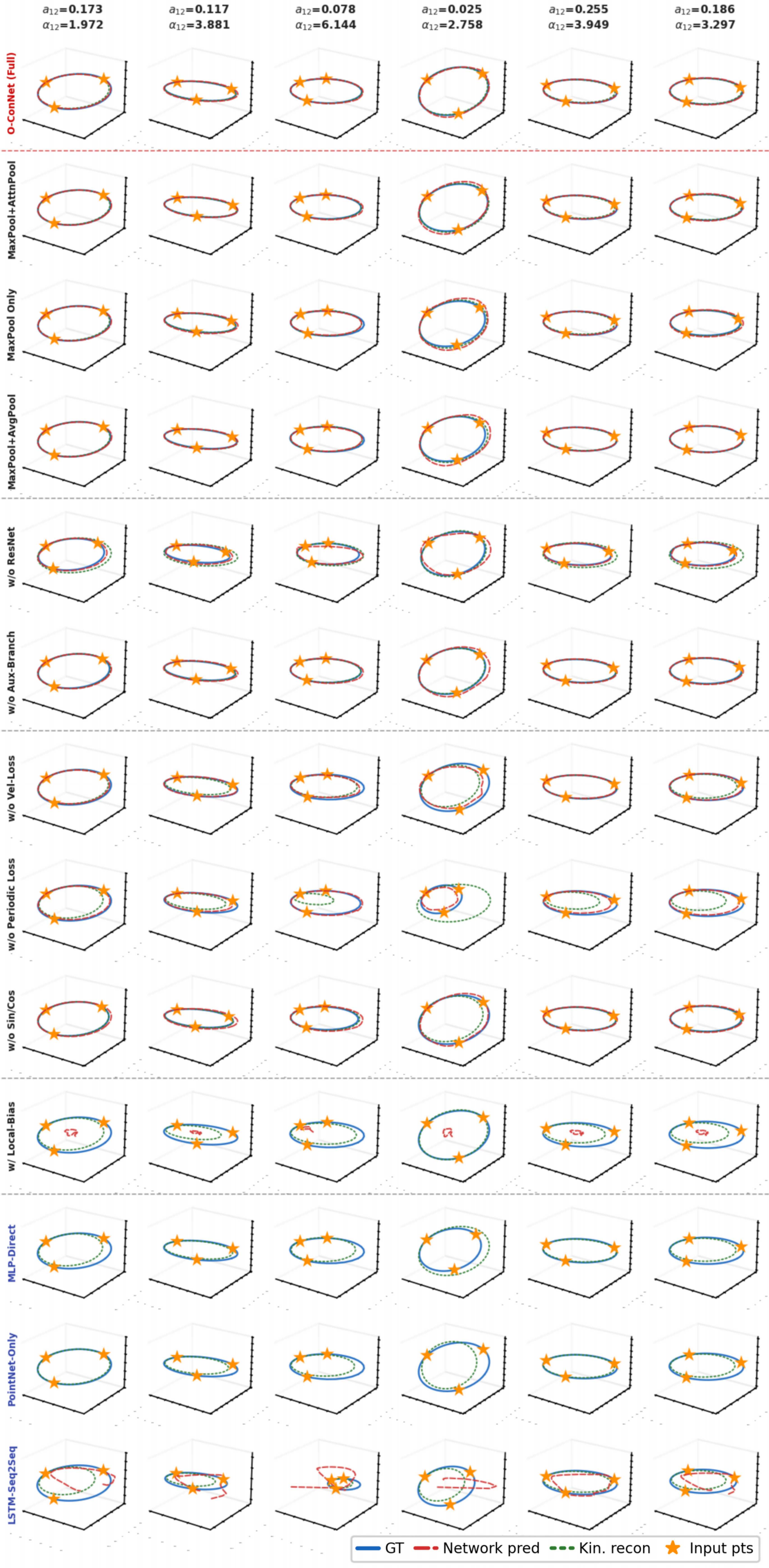}
  \caption{3D trajectory comparison of ablation configurations and baselines across randomly selected validation samples. Color coding follows Fig.~\ref{fig:network}. Baselines without a trajectory-decoding branch (MLP-Direct, PointNet-Only) predict parameters directly and thus show no Pred curve (red dashed).}
  \label{ablation_baseline_grid}
\end{figure}

\begin{table*}[!t]
\centering
\caption{Comprehensive performance summary on the validation set. MLP-Direct and PointNet-Only lack trajectory decoding branches ($^\dagger$); corresponding metrics are marked ``--''. Each ablation alters only one factor while keeping all others consistent with the full model. All metrics are computed in the normalized coordinate system; lower is better. All values are mean\,$\pm$\,std over 10 independent runs.}
\label{tab:all_results}
\setlength{\tabcolsep}{2pt}
\renewcommand{\arraystretch}{1.05}
\resizebox{\textwidth}{!}{%
\begin{tabular}{l|l|l|r|rrr|rr}
\toprule
& \textbf{Sub-group} & \textbf{Method} & \textbf{Params} & \textbf{Param-MAE}$\downarrow$ & $\mathbf{a_{12}}$\textbf{-MAE}$\downarrow$ & $\boldsymbol{\alpha_{12}}$\textbf{-MAE}$\downarrow$ & \textbf{Traj-MAE}$\downarrow$ & \textbf{Vel-MAE}$\downarrow$ \\
\midrule
Proposed & & \textbf{O-ConNet (Full)} & \textbf{5.00M} & $0.276{\pm}0.077$ & $\mathbf{0.0063{\pm}0.0013}$ & $\mathbf{0.316{\pm}0.063}$ & $\mathbf{0.145{\pm}0.018}$ & $\mathbf{0.017{\pm}0.002}$ \\
\midrule
\multirow{9}{*}{Ablation}
 & \multirow{3}{*}{Enc.\ Pooling} & MaxPool+AttnPool  & 5.13M & $\mathbf{0.260{\pm}0.084}$ & $0.0065{\pm}0.0008$ & $0.337{\pm}0.118$ & $0.149{\pm}0.025$ & $0.017{\pm}0.003$ \\
 & & MaxPool only       & 4.97M & $0.283{\pm}0.109$ & $0.0067{\pm}0.0010$ & $0.339{\pm}0.131$ & $0.175{\pm}0.031$ & $0.018{\pm}0.003$ \\
 & & MaxPool+AvgPool    & 5.10M & $0.286{\pm}0.114$ & $0.0065{\pm}0.0012$ & $0.336{\pm}0.122$ & $0.154{\pm}0.033$ & $0.018{\pm}0.004$ \\
\cmidrule{2-9}
 & \multirow{2}{*}{Param.\ Head} & w/o ResNet Residual   & 4.97M & $0.391{\pm}0.079$ & $0.0088{\pm}0.0018$ & $0.502{\pm}0.106$ & $0.199{\pm}0.017$ & $0.021{\pm}0.001$ \\
 & & w/o Aux.\ Branch Loss & 5.00M & $0.301{\pm}0.076$ & $0.0056{\pm}0.0009$ & $0.345{\pm}0.090$ & $0.160{\pm}0.025$ & $0.018{\pm}0.002$ \\
\cmidrule{2-9}
 & \multirow{3}{*}{Loss Func.} & w/o Velocity Loss          & 5.00M & $0.342{\pm}0.059$ & $0.0165{\pm}0.0022$ & $0.399{\pm}0.075$ & $0.185{\pm}0.013$ & $1.859{\pm}0.797$ \\
 & & w/o Periodic Angle Loss    & 5.00M & $0.425{\pm}0.036$ & $0.0340{\pm}0.0055$ & $0.781{\pm}0.075$ & $0.287{\pm}0.044$ & $0.031{\pm}0.003$ \\
 & & w/o sin/cos Supervision    & 5.00M & $0.422{\pm}0.041$ & $0.0150{\pm}0.0011$ & $0.539{\pm}0.071$ & $0.199{\pm}0.031$ & $0.023{\pm}0.002$ \\
\cmidrule{2-9}
 & Decoder & w/ Local Attn.\ Bias       & 5.14M & $0.520{\pm}0.131$ & $0.0139{\pm}0.0071$ & $0.650{\pm}0.149$ & $1.705{\pm}0.068$ & $0.183{\pm}0.006$ \\
\midrule
\multirow{3}{*}{Baselines}
 & & MLP-Direct$^\dagger$    & 0.11M & $0.331{\pm}0.047$ & $0.0141{\pm}0.0025$ & $0.550{\pm}0.043$ & --    & --    \\
 & & PointNet-Only$^\dagger$ & 0.48M & $0.554{\pm}0.103$ & $0.0161{\pm}0.0015$ & $0.719{\pm}0.066$ & --    & --    \\
 & & LSTM-Seq2Seq            & 3.25M & $0.790{\pm}0.152$ & $0.0234{\pm}0.0084$ & $0.875{\pm}0.043$ & $1.230{\pm}0.138$ & $0.181{\pm}0.032$ \\
\bottomrule
\end{tabular}}
\end{table*}

The normalized 42,860-sample dataset is partitioned at an 8:2 train/val ratio for all experiments. Training employs AdamW at a peak learning rate of $3\!\times\!10^{-4}$ with a short linear warm-up followed by cosine annealing, balancing early stability against broad mid-training exploration. Each configuration completes 100 epochs on an NVIDIA RTX~4090 with batch size 256 and mixed-precision~acceleration.

\subsection{Evaluation Metrics}
Five metrics are defined to cover all three output modalities: parameter prediction, trajectory reconstruction, and velocity reconstruction. All are computed in the normalized coordinate system, where lower values indicate better~performance.
Let the predicted parameters be $(\hat{a}_{12},\hat{\alpha}_{12})$ and the ground truth be $(a_{12},\alpha_{12})$; predicted and ground-truth trajectories/velocity fields are both in $\mathbb{R}^{T\times3}$ ($T{=}64$); $\langle\cdot\rangle$ denotes averaging over~samples.
The five metrics are defined~as:

\begin{align}
  a_{12}\text{-MAE}       &= \bigl\langle|\hat{a}_{12} - a_{12}|\bigr\rangle \label{eq:a12mae}\\[2pt]
  \alpha_{12}\text{-MAE}  &= \bigl\langle|\operatorname{wrap}(\hat{\alpha}_{12} - \alpha_{12})|\bigr\rangle, \notag\\
                          &\quad \operatorname{wrap}(\delta) \triangleq [(\delta+\pi)\bmod 2\pi]-\pi \label{eq:alpha12mae}\\[2pt]
  \text{Param-MAE}        &= \tfrac{1}{2}\bigl(a_{12}\text{-MAE} + \alpha_{12}\text{-MAE}\bigr) \label{eq:parammae}\\[2pt]
  \text{Traj-MAE}         &= \bigl\langle\tfrac{1}{Td}\|\hat{\mathbf{P}}-\mathbf{P}\|_1\bigr\rangle \label{eq:trajmae}\\[2pt]
  \text{Vel-MAE}          &= \bigl\langle\tfrac{1}{Td}\|\hat{\mathbf{V}}-\mathbf{V}\|_1\bigr\rangle \label{eq:velmae}
\end{align}

$\operatorname{wrap}(\cdot)$ folds the angular difference into $(-\pi,\pi]$, eliminating the systematic overestimation caused by the $2\pi$ periodic boundary on naive MAE. Param-MAE takes the arithmetic mean of the two parameter errors and serves as the composite ranking~metric.

To cover comparisons across different architectural paradigms, three baseline groups are established. \textbf{MLP-Direct} flattens the 3 input points and maps them to 2-dimensional parameter output via a 4-layer fully connected network, without a trajectory decoding~branch.
\textbf{PointNet-Only} retains the point cloud encoder and parameter prediction MLP but removes the Transformer decoder and ResNet residual structure, regressing parameters directly from the 512-dimensional latent~vector.
\textbf{LSTM-Seq2Seq} employs a two-layer LSTM encoder--decoder architecture to generate trajectories in parallel, then extracts parameters from the reconstructed trajectory via PointNet; it is the only baseline with trajectory output~capability.

\subsection{Comparative Experiments}\label{sec:main_results}

Table~\ref{tab:all_results} reports results for all~methods.
Single-path baselines reveal how much the trajectory reconstruction branch contributes. MLP-Direct proves that direct regression from sparse input is feasible (Param-MAE $0.331\pm0.047$), yet without any trajectory output its predictions have no kinematic self-consistency guarantee. PointNet-Only adds stronger feature extraction but also relies on a single path; starved of the auxiliary gradient that trajectory reconstruction would supply, it actually regresses to Param-MAE $0.554\pm0.103$, worse than the shallower MLP-Direct. LSTM-Seq2Seq, the only baseline with trajectory output, is undermined by autoregressive error accumulation, reaching Param-MAE $0.790\pm0.152$ and Traj-MAE $1.230\pm0.138$. O-ConNet reduces those two figures to $0.145\pm0.018$ and $0.017\pm0.002$, gains of 88.2\% and 90.6\% over~LSTM-Seq2Seq.

Fig.~\ref{fig:traj_vis} illustrates the complementary nature of the two output streams. Pred closely follows the GT envelope across all projection planes, confirming that the decoder has internalized the spatial geometry of the closed-loop trajectory. The independently computed Kin curve converges to the same shape via a completely separate path, providing a physical sanity check: agreement between Kin and GT validates that the predicted parameters encode genuine kinematic structure, not merely a plausible-looking spatial~curve.

\subsection{Ablation Study}\label{sec:ablation}

Each ablation changes one design factor while keeping the rest identical to the full model; Fig.~\ref{ablation_baseline_grid} provides trajectory-level visual evidence alongside the quantitative~table.

On pooling, both MaxPool+AttnPool and AttnPool-only were evaluated. MaxPool+AttnPool achieves a slight advantage on Param-MAE ($0.260\pm0.084$ vs.\ $0.276\pm0.077$), whereas AttnPool-only leads on Traj-MAE ($0.145$ vs.\ $0.149$) and yields more balanced performance across all metrics. We therefore adopt AttnPool as the final encoder despite its marginal Param-MAE deficit. MaxPool-only and MaxPool+AvgPool both trail the attention-pooling family, confirming that learned attention weighting is the decisive~element.

Removing the ResNet residual structure (w/o\,ResNet) raises Param-MAE from 0.276 to 0.391. A telling dissociation appears in the trajectory visualizations: the Pred curve still approximately follows the GT shape, while the independently computed Kin curve drifts noticeably away. This gap between visual plausibility and analytical precision confirms that the residual branch is indispensable for the parameter head to achieve the fine-grained accuracy that trajectory resemblance alone cannot guarantee. Removing the auxiliary branch loss has a comparatively milder~effect.

Three loss-function ablations reveal qualitatively distinct failure modes. Removing the velocity loss (w/o\,Vel-Loss) preserves the coarse trajectory shape but drives Vel-MAE from 0.017 to 1.859, and the distorted velocity features subsequently propagate noisy gradients into the parameter head, quietly eroding parameter accuracy. Omitting the periodicity-aware angle loss (w/o\,Periodic) nearly triples $\alpha_{12}$-MAE from 0.316 to 0.781, with visible shape distortions concentrated near twist angles of $0$ or $\pi$ where the $2\pi$ wrap-around is most disruptive. Dropping sin/cos supervision (w/o\,Sin/Cos) raises $\alpha_{12}$-MAE to 0.539, confirming that direct geometric supervision on the angular representation is comparably~critical.

Local-Bias inflicts the worst single degradation, pushing Traj-MAE from 0.145 to 1.705. Because the Gaussian log-space bias is active only during training, the decoder learns an attention distribution it cannot replicate at inference; compounding this inconsistency, any fixed locality prior prevents the global cross-position attention that coherent curve extrapolation from three sparse points~demands.

Throughout all configurations $a_{12}$-MAE remains well below $\alpha_{12}$-MAE (0.005--0.034 vs.\ 0.316--0.875), identifying twist-angle estimation as the persistent bottleneck. Even modest $\Delta\alpha_{12}$ errors are nonlinearly amplified by the Bennett closure constraint, which explains why Kin curves can diverge sharply from GT in samples where Pred still looks~reasonable.

\section{CONCLUSION}\label{sec:conclusion}

We have proposed O-ConNet, an end-to-end framework that infers Bennett mechanism parameters and reconstructs full motion trajectories from only three sparse waypoints, implicitly encoding closed-loop geometric constraints through data rather than explicit~equations.
On the validation split, $a_{12}$-MAE has reached $0.0063$, $\alpha_{12}$-MAE has reached $0.316$, and Traj-MAE has reached $0.145$, with the twist-angle term remaining the dominant source of error due to the nonlinear amplification imposed by the Bennett closure~constraint.
Future work will extend data generation and evaluation to other over-constrained families such as Bricard and Goldberg, and explore whether the periodicity-aware angular representation and dual-path parameter head transfer without architectural~modification.

\addtolength{\textheight}{-12cm}

\bibliographystyle{IEEEtran}
\bibliography{ref}

\end{document}